\documentclass[%
 reprint,
 amsmath,amssymb,
 aps,
superscriptaddress]{revtex4-1}

\usepackage{graphicx}
\usepackage{dcolumn}
\usepackage{comment}
\usepackage{bm}
\usepackage{color}
\usepackage{hyperref}

\begin{document}

\preprint{APS/123-QED}

\title{Bellybutton: Accessible and Customizable Deep-Learning Image Segmentation }

\author{Sam Dillavou}
 \email{dillavou@sas.upenn.edu}

\author{Jesse M. Hanlan}

\author{Anthony T. Chieco}

\author{Hongyi Xiao}
\affiliation{%
 Department of Physics and Astronomy, University of Pennsylvania, Philadelphia, PA, 19104
}
\author{Sage Fulco}
\affiliation{%
 Department of Mechanical Engineering and Applied Mechanics, University of Pennsylvania, Philadelphia, PA, 19104
}
\author{Kevin T. Turner}
\affiliation{%
 Department of Mechanical Engineering and Applied Mechanics, University of Pennsylvania, Philadelphia, PA, 19104
}
\author{Douglas J. Durian}
 \affiliation{%
 Department of Physics and Astronomy, University of Pennsylvania, Philadelphia, PA, 19104
}
\affiliation{Center for Computational Biology, Flatiron Institute, Simons Foundation, New York, NY 10010, USA}

\date{\today}

\begin{abstract}
The conversion of raw images into quantifiable data can be a major hurdle in experimental research, and typically involves identifying region(s) of interest, a process known as segmentation. Machine learning tools for image segmentation are often specific to a set of tasks, such as tracking cells, or require substantial compute or coding knowledge to train and use. Here we introduce an easy-to-use (no coding required), image segmentation method, using a 15-layer convolutional neural network that can be trained on a laptop: Bellybutton. The algorithm trains on user-provided segmentation of example images, but, as we show, just one or even a portion of one training image can be sufficient in some cases. We detail the machine learning method and give three use cases where Bellybutton correctly segments images despite substantial lighting, shape, size, focus, and/or structure variation across the regions(s) of interest. Instructions for easy download and use, with further details and the datasets used in this paper are available at \href{http://www.pypi.org/project/Bellybuttonseg}{pypi.org/project/Bellybuttonseg}.
\end{abstract}

\maketitle


\section{Introduction}

\begin{figure*} 
\includegraphics[width = 17.4cm]{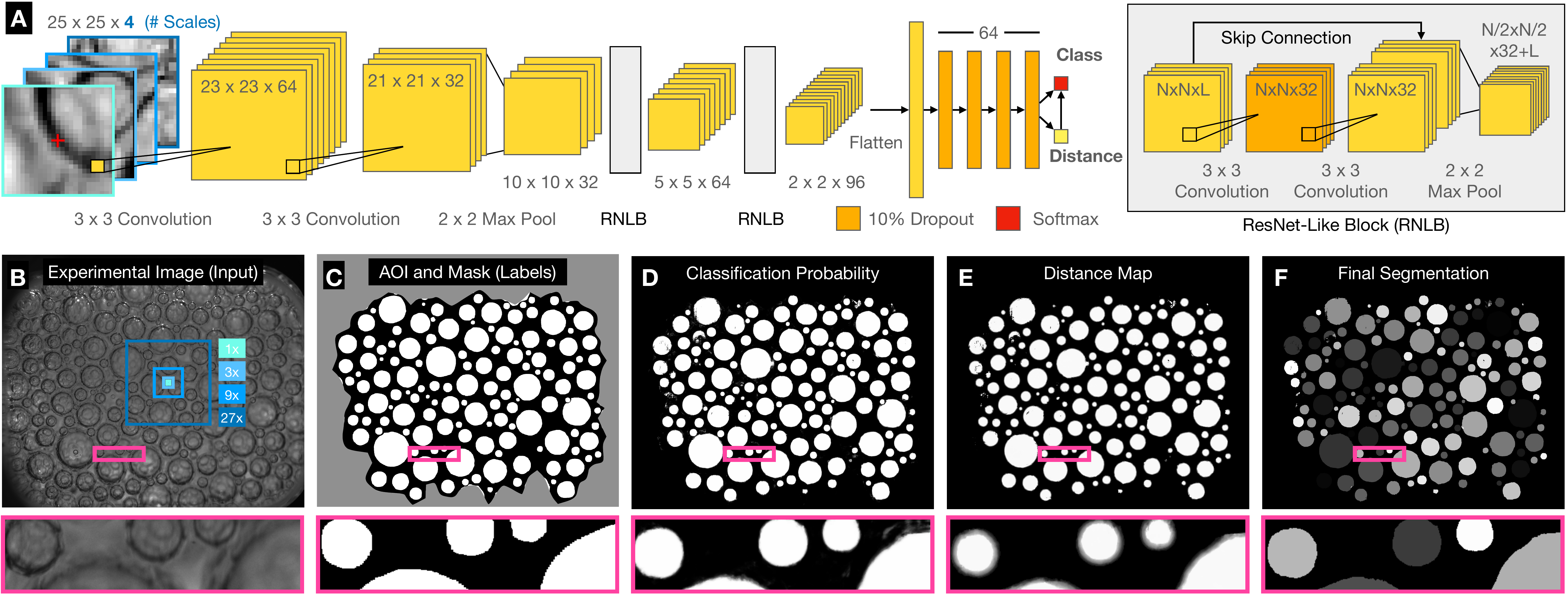}
\caption{\textbf{The Bellybutton Method} 
\textbf{(A)} Architecture of the 15-layer convolutional neural network. Multiple scales of an experimental image, each reduced to 25x25 pixels, are simultaneously taken as a single input. The network consists of two 3x3 convolutional layers followed by a 2x2 max pooling layer. This pattern is repeated twice more, each with skip connections as shown. The final 2x2x96 layer is flattened, fed through four dense layers and produces a two output scalars, one signifying the class of the pixel (inside or outside of a region), the other the distance to the nearest region edge. \textbf{(B)} An example experimental image, overlaid with the chosen input scales 1, 3, 9, and 27x. \textbf{(C)} User-defined mask, in this case binary as no segments are in contact. User may also define an area of Interest (AOI), which in this example removes the edges of the image (gray) from training. \textbf{(D)} Class probability output after training. The network generates a prediction score on a pixel-by-pixel basis. \textbf{(E)} Distance map to outside of a particle. Values are capped at a user-specified value, in this case 10 pixels, so much of the image appears binary. The zoomed-in region highlights the gray-scale output near the edges of the bubbles. \textbf{(F)} Final segementation is produced by watershedding the binarized classification probability (D) using the distance map (E). (D) and (E) are also saved if desired.
}
\label{fig1}
\end{figure*} 

Extracting quantitative information from image data is a major step in many fields of research. Prior to the last decade, state of the art algorithms typically focused on highly specific use cases, such as tracking spherical particles \cite{crocker_methods_1996} or identifying astronomical light sources \cite{bertin_sextractor_1996}. These algorithms were typically task specific - aiming to identify predefined features - as opposed to machine learning algorithms that are more adaptive. In fact, reviews as late as 2015 did not even mention machine learning (ML) \cite{manzo_review_2015}. Progress is still being made in this domain today \cite{yucel_toolbox_2021}. Since the introduction of AlexNet \cite{krizhevsky_imagenet_2012} in 2012, the capacity of ML methods in this arena has moved at a breathtaking pace, fueled largely by the success of convolutional neural networks  (CNNs) \cite{chai_deep_2021}. This class of techniques allows a more general approach to quantification of image data, including addressing more nuanced and harder-to-formulate questions by requiring only correct examples as training data. More specifically, the task of segmenting an image - identifying the pixels that comprise one or more objects or regions of interest - has become a large focus \cite{minaee_image_2021}, as it allows researchers to rapidly and deeply analyze complex data. While state-of-the-art benchmarks in this domain \footnote{https://mlcommons.org/en/} require enormous computation and are thus out of even a skilled single user's reach, software tools like Keras \cite{chollet2015keras}, an Application Program Interface (API) for Python, greatly simplify the process of creating smaller, custom neural network solutions, in principle in just a few lines of code. However, in practice the process is rarely that simple, and for those unfamiliar with deep neural networks, many pieces of the process become daunting; optimizing the many user-defined ``hyper-parameters" of the algorithm, picking the right network, cleaning the data, and possibly learning a new programming language can each require a lot of additional effort. 

As a result, a large and recent body of work has been focused on methods and software packages for simplifying this process. The majority focused on biological research, specifically the tracking of cells from microscopy data \cite{midtvedt_quantitative_2021, ershov_trackmate_2022, ulman_objective_2017, stringer_cellpose_2021, zheng_strains_2022,newby_convolutional_2018,berg_ilastik_2019,nguyen_automatically_2017}, but similar works tackle goals ranging from identifying and tracking 2D materials like graphene \cite{masubuchi_deeplearningbased_2020} to segmenting other medical or biological imaging data \cite{malhotra_deep_2022, ronneberger_unet_2015, ciresan_deep_2012,haertter_deepprojection_2022}, images of flora and fauna \cite{niedballa_imageseg_2022}, scanning electron microscopy images for material science \cite{ruhle_workflow_2021, azimi_advanced_2018}, astronomical data \cite{ostdiek_image_2022, hausen_morpheus_2020}, particle physics \cite{li_reconstructing_2021}, and more. Typically these works compete for highest accuracy on benchmark data sets \cite{ulman_objective_2017}, or ease of use for pre-specified domains (very often biological data) \cite{midtvedt_quantitative_2021,ershov_trackmate_2022}. While many of these methods are likely applicable for tasks outside of their intended application, e.g. \cite{berg_ilastik_2019}, few are explicitly designed for general use. 

\begin{figure*} 
\includegraphics[width = 17.4cm]{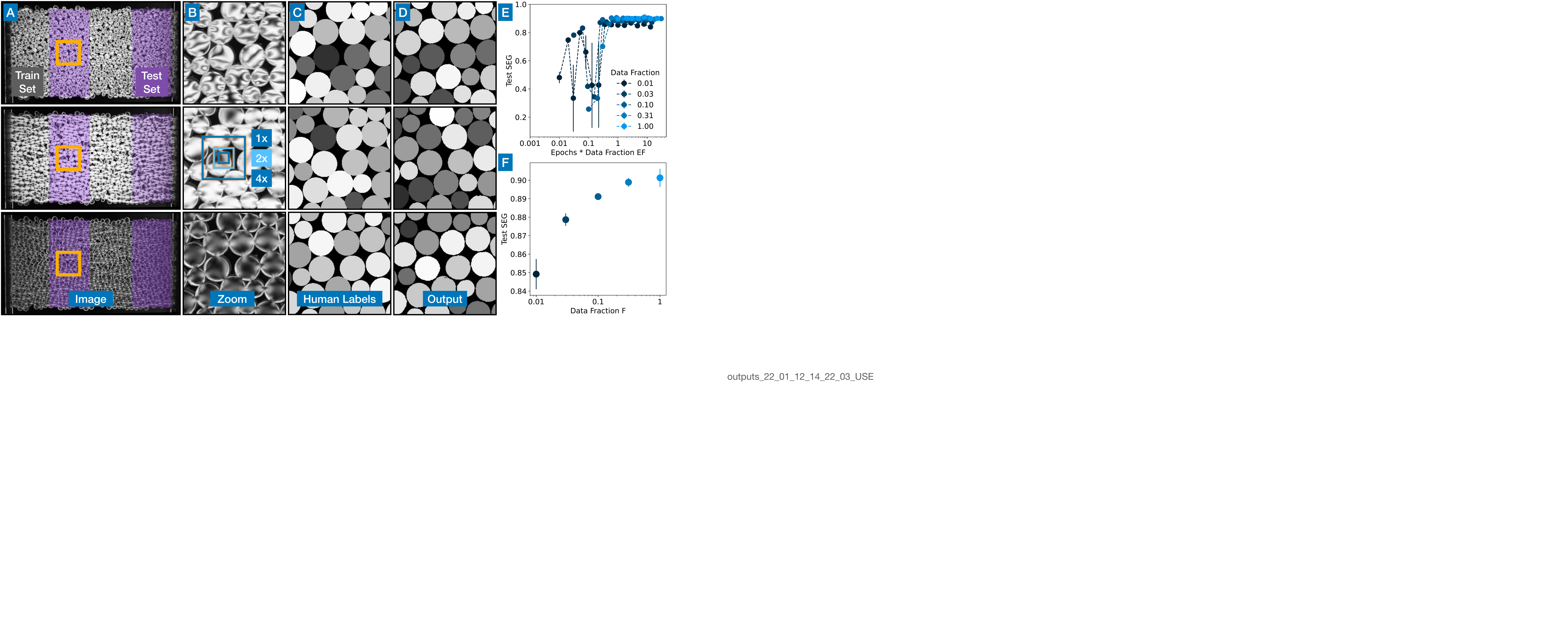}
\caption{\textbf{3D Printed Photoelastic Disks} \textbf{(A)} Images of a 3D printed photoelastic material in the shape of a granular packing under three stress states (high, medium, low). Each was divided into four sections, two of which (gray) were used for training, and two (purple) were used for evaluation. A single network was trained using all six training regions, and tested on all six test regions. \textbf{(B)} Zoom in on orange-framed region in (A). Note the variety of lighting patterns on each disk. Teal and blue superimposed squares are the image scales fed into the network for this task. \textbf{(C)} User-generated masks for these zoomed in regions (which are part of the test set). \textbf{(D)} Final segmentation output for the zoomed in region. Note that the colors serve to differentiate regions; there is no attempt to match the colors between (C) and (D). \textbf{(E)} SEG score for the test set as a function of Epochs times Training Fraction $EF$. Training fraction $F$ is denoted by color, and is the portion of the training data used in training the network, with each data point shown to the network once per epoch. SEG score is an indicator of segmentation quality, and is calculated by dividing the intersection of generated regions and their corresponding true regions with their union, and averaging for all true regions (see text for further explanation).  \textbf{(F)} SEG score for all runs with $EF\geq 3 $ as a function of data fraction $F$. Note the diminishing returns on this task for high $F$.
}
\label{fig2}
\end{figure*} 
 
Here we introduce an easy-to-use segmentation solution aimed at a broad array of research applications, named ``Bellybutton." Bellybutton uses a 15-layer convolutional neural network that can be trained on as little as one (or a portion of one) image with user-defined segmentation, and can account for variations in size, lighting, rotation, focus, or shape of desired segmentation regions, as is common in research applications. The algorithm operates on a pixel-by-pixel basis, determining if each is inside or outside of a segmentation (`innies' or `outies,' hence the name Bellybutton). The algorithm can analyze input images of varying shape and size, and automatically performs a variety of data augmentation, including flipping and rotating images, normalizing brightness across images, and evenly sampling innies and outies.  Bellybutton requires no coding knowledge, and can be trained and run on a laptop. We detail its performance and flexibility through several use cases including segmenting bubbles with poor lighting and focus, semi-transparent, tightly packed particles that have intricate birefringence patterns, and tracking a thin clear lattice of material that fractures over time. Each of these data sets is available online, along with a guide for Bellybutton's use on new data sets. 

\section{Method}
Bellybutton operates on a pixel-by-pixel basis, scanning images and using the neighborhood around a given point in an image to determine if a pixel is inside or outside of a segment, as well as how far from that segment's edge. It uses a deep convolutional neural network (CNN), whose structure is shown schematically in Fig.~\ref{fig1}A. The CNN consists of 3x3 convolutional layers, 2x2 max pooling layers, skip connections inspired by ResNet \cite{he_deep_2015}, and ends with four dense layers feeding into two outputs - a classification of pixel type (inside or outside a region), and a distance-from-region-edge scalar value, which is used to separate distinct regions in contact. The scalar value is trained to vary between 0 (for all outside pixels) to a maximum value set by the user (typically 10), allowing the system to localize region edges while easily satisfying this output when it is unimportant, for example in the center of a 100 pixel-wide region. The chosen network architecture strikes a balance between being small enough to train rapidly from scratch on a laptop, while being large enough to generate valid segmentation on nontrivial problems. The choice of a CNN has been the standard for segmentation problems \cite{hausen_morpheus_2020,chai_deep_2021,azimi_advanced_2018,chai_deep_2021,stringer_cellpose_2021,newby_convolutional_2018,haertter_deepprojection_2022,malhotra_deep_2022,ronneberger_unet_2015,niedballa_imageseg_2022,ruhle_workflow_2021, ostdiek_image_2022}, as it allows the network natural access to spatial information. The decreasing layer size is also standard, and gives the network sufficient flexibility to hierarchically analyze spatial patterns without superfluous parameters. The network itself takes multiple size subsets of an image as input, centered around the pixel in question, each down-sampled to 25x25 pixels. This sampling process is performed automatically during training and prediction, and gives the network the ability to analyze multiple length scales while keeping input size minimal. A typical example is shown in Fig.~\ref{fig1}A and B using 1, 3, 9, and 27x scales.

For training, a user may provide individually-labeled segmentation maps, that is, every pixel in a particular segment must contain the same number, unique to that segment. Alternatively, if no segments are in contact, a user-provided binary mask is sufficient. Pixels are each then given a classification label that corresponds to `innie' (inside a segmented region' or `outie.' Optionally the user may exclude regions of an image using a binary Area of Interest (AOI) mask, as indicated by the excluded gray area in Fig.~\ref{fig1}C. The distance to segment edges is also calculated from this mask, and used to train the scalar output.

To avoid prolonged training, the user may select to train using a fraction of available training data. We find that near optimal results are often reached without using all available pixels (see Fig.~\ref{fig2}E.) Furthermore, rotated and flipped images are (optionally) used in training to prevent overfitting. Once trained, Bellybutton produces a score of 0 (outside) to 1 (inside a region) for each pixel, shown in Fig.~\ref{fig1}D, which is binarized to produce an innie-vs-outie map. Finally, the output of the scalar distance-to-region-edge, shown in Fig.~\ref{fig1}E, is used to watershed the `innie' pixels into distinct regions to produce a segmented map, as in Fig.~\ref{fig1}E. Data used in this figure, aqueous foams in microgravity, comes from Ref.~\cite{pasquet_aqueous_2023}, which was the first work to utilize Bellybutton.

\begin{figure*} 
\includegraphics[width = 17.4cm]{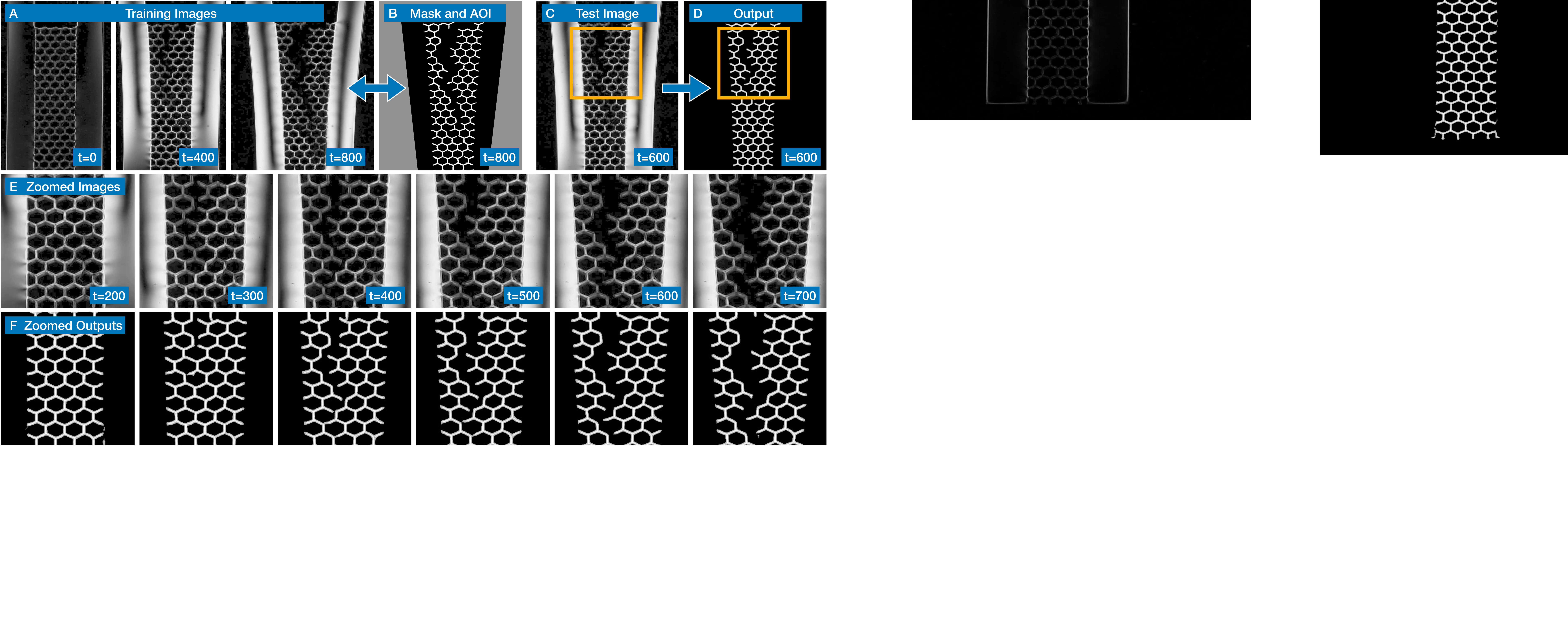}
\caption{\textbf{Tracking a Changing Structure with Bellybutton} \textbf{(A)} Training images of a fracturing lattice. Image contrast and brightness have been enhanced, and the top 2/3 of each image is shown. Note that these are the only training images, but that we have spread them out in time to encompass a wide range of situations. \textbf{(B)} Binary mask for the third training image with superimposed area of interest (gray). \textbf{(C)} Example test image and \textbf{(D)} accompanying Bellybutton-generated distance map output. Orange square denotes location of zoomed regions in (E) and (F). \textbf{(E)} Zoomed in (enhanced) images with \textbf{(F)} corresponding Bellybutton-generated distance map for many time steps.
}
\label{fig3}
\end{figure*} 

\section{Example Uses}

Bellybutton is effective for a variety of purposes. Here we use the example of segementing a 3D printed photoelastic material in the shape of a granular packing. This material is illuminated between cross-polarizers such that it develops a birefringence pattern when under mechanical stress. This lighting is useful experimentally, but complicates the tracking process; previous experiments using photoelastic granular disks have required two sets of images, one with regular lighting to track particles, and second one with the birefringence pattern to analyze force \cite{daniels_photoelastic_2017}. Bellybutton was trained on two fourths of three images of this system, under low, medium, and high stress, and tested on the remaining two fourths of each image, shaded purple in Fig.~\ref{fig2}A. While remaining roughly the same shape, the particles present a wide variety of patterns as the stress changes. Furthermore, a variety of confounding factors make this segmentation more difficult: A substantial portion of the image (the left and right edges) is out of focus. The camera is close enough to the sample that only particles in the center are imaged head-on, leading to different viewing angles for particles near the edges of the system. Finally, particles near the left and right edge are tilted sufficiently such that their edges are exposed to the camera.

The input scales used are shown in Fig.~\ref{fig2}B, overlaid on zoomed-in data. Segmentation is successful, with the majority of errors concentrated at the bottom of the leftmost image, where contrast and focus are worst. Typical regions are successfully segmented, as seen by comparing Fig.~\ref{fig2}C and D, taken from the test set. 

For quantitative analysis of these results, we utilize the SEG score from Ref.~\cite{ulman_objective_2017}, which compares each true region with the identified region of highest overlap. We find this metric to be the most indicative of performance by eye, although many others are commonly used \cite{ulman_objective_2017, minaee_image_2021}. For each true region $R_i$, a `Jaccard index' is calculated with the Bellybutton-generated region $B_i$ of highest overlap, by dividing the area of their intersection by the area of their union. True regions that do not have an intersection of at least one half of their area are given a score of 0. The SEG reported is the average of all such scores for a given dataset, with a perfect score being 1. A detailed explanation of the calculation is given in \footnote{https://public.celltrackingchallenge.net/documents/SEG.pdf}. Bellybutton was reliably able to beat a 0.9 SEG score on the test set for this data.

In the highlighted example the entire training set was used, and the network was trained for $E=2$ epochs (each training data point was shown to the network twice). For practical use however, it may not be necessary to use even this much data (half of three images), as shown in Fig.~\ref{fig2}E. A sub-sampling option is given as a parameter in the Bellybutton package, named `fraction.' This value indicates the fraction (0-1] of available training pixels that the algorithm will use to train the neural network. For values below 1, individual pixels are randomly chosen, but at a rate that ensures that innies and outies are equally represented \footnote{This can also be modified easily via the parameters of the algorithm, to instead represent innies and outies in the ratio they are present in the images.}. We find that accuracy for a variety of problems is dependent on the quantity
\begin{equation}
    EF = T/M
\end{equation}
being sufficiently high, where $E$ is the number of epochs in training, $M$ is the size of the total training set, $F$ is the fraction of the training set that is used, and $T=EFM$ is the total number of training steps. This dependency is shown by the data collapse in Fig.~\ref{fig2}E. As a result, smaller data fractions $F$ can be used to suss out the tractability of a problem. In this example, even tiny fractions of the training data can still yield passable results, as seen by the modest dependence of SEG on data fraction in Fig.~\ref{fig2}F, however for optimal results, a larger fraction of the data must be used, to give the network access to a wider variety of examples. Overall, more data is typically better, but we often find that $F\geq 0.1$ gives reasonable results for systems with many repeated particles, like the one shown in Fig.~\ref{fig2}. An important caveat is that these training data should be taken from a sufficiently varied set of images and locations within those images to encompass the range of the desired data set.

Bellybutton is also useful for structure-finding. In the following example a lattice of laser-cut acrylic (Polymethyl methacrylate or PMMA) is slowly fractured while lit between cross-polarizers to reveal changes in internal stress. These changes to the material's structure as well as its brightness, shown in Fig.~\ref{fig3}A and E, make it very difficult to track algorithmically. Using just three training images with human-generated masks, Bellybutton is capable of tracking the fracturing structure through time, as shown in Fig.~\ref{fig3}E and F, despite lighting and focus changes. The package includes options for a binarized output, or a distance-to-edge output, which is shown here. The latter can be helpful for skeletonizing a structure, and to suppress noise and error.

\section{How and When to Use Bellybutton}

We have tried to make Bellybutton as accessible as possible. It is downloadable as a python package, which can be easily installed with one command, and utilizing Bellybutton requires no coding. Instructions for use, details for how to customize training and hyper-parameters, and much more can be found at \href{http://www.pypi.org/project/Bellybuttonseg}{pypi.org/project/Bellybuttonseg}. Starting a project is as simple as running a single command, and Bellybutton creates a folder structure to add images, masks, and areas of interest. Adjusting the parameters of training and testing are done through editing an automatically-generated text file. Furthermore, we have provided the data sets used in each figure as example projects that can be downloaded in one command, set up, and run on a laptop. Deploying one of these example projects takes under a minute, plus training time (computer dependent).

While only three examples of Bellybutton's potential uses are shown, its flexibility should make it useful in a wide variety of situations. Regions are not limited to single particles; masks might specify the two connected regions of a dimer, or a disk and a mark on its surface indicating its rotational position as separate regions, allowing them both to be segmented simultaneously. The same approach could be applied to a cell and its nucleus, an insect and its head or feet, a particle \textit{and its previous position}, allowing velocity to be approximated from single images. Regions can be used to identify particle classes as well; segmenting only particles of a given shape, size, or orientation will prompt Bellybutton to do the same. A broad rule of thumb is if a region is easily identifiable by eye, it is a good candidate for Bellybutton. This class of image segmentation problems is both frustrating and common in research, and we believe giving users an easy-to-use but flexible method like Bellybutton will save countless hours in the lab.

\acknowledgements
We thank Marina Pasquet for assistance with ISS data from the Foam-C project of ESA \cite{pasquet_aqueous_2023}, and Kieran A Murphy for helpful discussions. This work was supported by NASA grant 80NSSC21K0898 and by NSF grant MRSEC/DMR-1720530.

\bibliography{bib}

\begin{thebibliography}{33}%
\makeatletter
\providecommand \@ifxundefined [1]{%
 \@ifx{#1\undefined}
}%
\providecommand \@ifnum [1]{%
 \ifnum #1\expandafter \@firstoftwo
 \else \expandafter \@secondoftwo
 \fi
}%
\providecommand \@ifx [1]{%
 \ifx #1\expandafter \@firstoftwo
 \else \expandafter \@secondoftwo
 \fi
}%
\providecommand \natexlab [1]{#1}%
\providecommand \enquote  [1]{``#1''}%
\providecommand \bibnamefont  [1]{#1}%
\providecommand \bibfnamefont [1]{#1}%
\providecommand \citenamefont [1]{#1}%
\providecommand \href@noop [0]{\@secondoftwo}%
\providecommand \href [0]{\begingroup \@sanitize@url \@href}%
\providecommand \@href[1]{\@@startlink{#1}\@@href}%
\providecommand \@@href[1]{\endgroup#1\@@endlink}%
\providecommand \@sanitize@url [0]{\catcode `\\12\catcode `\$12\catcode
  `\&12\catcode `\#12\catcode `\^12\catcode `\_12\catcode `\%12\relax}%
\providecommand \@@startlink[1]{}%
\providecommand \@@endlink[0]{}%
\providecommand \url  [0]{\begingroup\@sanitize@url \@url }%
\providecommand \@url [1]{\endgroup\@href {#1}{\urlprefix }}%
\providecommand \urlprefix  [0]{URL }%
\providecommand \Eprint [0]{\href }%
\providecommand \doibase [0]{https://doi.org/}%
\providecommand \selectlanguage [0]{\@gobble}%
\providecommand \bibinfo  [0]{\@secondoftwo}%
\providecommand \bibfield  [0]{\@secondoftwo}%
\providecommand \translation [1]{[#1]}%
\providecommand \BibitemOpen [0]{}%
\providecommand \bibitemStop [0]{}%
\providecommand \bibitemNoStop [0]{.\EOS\space}%
\providecommand \EOS [0]{\spacefactor3000\relax}%
\providecommand \BibitemShut  [1]{\csname bibitem#1\endcsname}%
\let\auto@bib@innerbib\@empty
\bibitem [{\citenamefont {Crocker}\ and\ \citenamefont
  {Grier}(1996)}]{crocker_methods_1996}%
  \BibitemOpen
  \bibfield  {author} {\bibinfo {author} {\bibfnamefont {J.~C.}\ \bibnamefont
  {Crocker}}\ and\ \bibinfo {author} {\bibfnamefont {D.~G.}\ \bibnamefont
  {Grier}},\ }\bibfield  {title} {\bibinfo {title} {Methods of {{Digital Video
  Microscopy}} for {{Colloidal Studies}}},\ }\href
  {https://doi.org/10.1006/jcis.1996.0217} {\bibfield  {journal} {\bibinfo
  {journal} {Journal of Colloid and Interface Science}\ }\textbf {\bibinfo
  {volume} {179}},\ \bibinfo {pages} {298} (\bibinfo {year}
  {1996})}\BibitemShut {NoStop}%
\bibitem [{\citenamefont {Bertin}\ and\ \citenamefont
  {Arnouts}(1996)}]{bertin_sextractor_1996}%
  \BibitemOpen
  \bibfield  {author} {\bibinfo {author} {\bibfnamefont {E.}~\bibnamefont
  {Bertin}}\ and\ \bibinfo {author} {\bibfnamefont {S.}~\bibnamefont
  {Arnouts}},\ }\bibfield  {title} {\bibinfo {title} {{{SExtractor}}:
  {{Software}} for source extraction.},\ }\href
  {https://doi.org/10.1051/aas:1996164} {\bibfield  {journal} {\bibinfo
  {journal} {Astronomy and Astrophysics Supplement Series}\ }\textbf {\bibinfo
  {volume} {117}},\ \bibinfo {pages} {393} (\bibinfo {year}
  {1996})}\BibitemShut {NoStop}%
\bibitem [{\citenamefont {Manzo}\ and\ \citenamefont
  {{Garcia-Parajo}}(2015)}]{manzo_review_2015}%
  \BibitemOpen
  \bibfield  {author} {\bibinfo {author} {\bibfnamefont {C.}~\bibnamefont
  {Manzo}}\ and\ \bibinfo {author} {\bibfnamefont {M.~F.}\ \bibnamefont
  {{Garcia-Parajo}}},\ }\bibfield  {title} {\bibinfo {title} {A review of
  progress in single particle tracking: From methods to biophysical insights},\
  }\href {https://doi.org/10.1088/0034-4885/78/12/124601} {\bibfield  {journal}
  {\bibinfo  {journal} {Reports on Progress in Physics}\ }\textbf {\bibinfo
  {volume} {78}},\ \bibinfo {pages} {124601} (\bibinfo {year}
  {2015})}\BibitemShut {NoStop}%
\bibitem [{\citenamefont {Y{\"u}cel}\ and\ \citenamefont
  {Velu}(2021)}]{yucel_toolbox_2021}%
  \BibitemOpen
  \bibfield  {author} {\bibinfo {author} {\bibfnamefont {H.}~\bibnamefont
  {Y{\"u}cel}}\ and\ \bibinfo {author} {\bibfnamefont {S.~K.~P.}\ \bibnamefont
  {Velu}},\ }\bibfield  {title} {\bibinfo {title} {Toolbox for tracking and
  analyzing crowded mixture of colloidal particles},\ }\href
  {https://doi.org/10.1016/j.colcom.2021.100546} {\bibfield  {journal}
  {\bibinfo  {journal} {Colloid and Interface Science Communications}\ }\textbf
  {\bibinfo {volume} {45}},\ \bibinfo {pages} {100546} (\bibinfo {year}
  {2021})}\BibitemShut {NoStop}%
\bibitem [{\citenamefont {Krizhevsky}\ \emph {et~al.}(2012)\citenamefont
  {Krizhevsky}, \citenamefont {Sutskever},\ and\ \citenamefont
  {Hinton}}]{krizhevsky_imagenet_2012}%
  \BibitemOpen
  \bibfield  {author} {\bibinfo {author} {\bibfnamefont {A.}~\bibnamefont
  {Krizhevsky}}, \bibinfo {author} {\bibfnamefont {I.}~\bibnamefont
  {Sutskever}},\ and\ \bibinfo {author} {\bibfnamefont {G.~E.}\ \bibnamefont
  {Hinton}},\ }\bibfield  {title} {\bibinfo {title} {{{ImageNet
  Classification}} with {{Deep Convolutional Neural Networks}}},\ }in\
  \href@noop {} {\emph {\bibinfo {booktitle} {Advances in {{Neural Information
  Processing Systems}}}}},\ Vol.~\bibinfo {volume} {25}\ (\bibinfo  {publisher}
  {{Curran Associates, Inc.}},\ \bibinfo {year} {2012})\BibitemShut {NoStop}%
\bibitem [{\citenamefont {Chai}\ \emph {et~al.}(2021)\citenamefont {Chai},
  \citenamefont {Zeng}, \citenamefont {Li},\ and\ \citenamefont
  {Ngai}}]{chai_deep_2021}%
  \BibitemOpen
  \bibfield  {author} {\bibinfo {author} {\bibfnamefont {J.}~\bibnamefont
  {Chai}}, \bibinfo {author} {\bibfnamefont {H.}~\bibnamefont {Zeng}}, \bibinfo
  {author} {\bibfnamefont {A.}~\bibnamefont {Li}},\ and\ \bibinfo {author}
  {\bibfnamefont {E.~W.~T.}\ \bibnamefont {Ngai}},\ }\bibfield  {title}
  {\bibinfo {title} {Deep learning in computer vision: {{A}} critical review of
  emerging techniques and application scenarios},\ }\href
  {https://doi.org/10.1016/j.mlwa.2021.100134} {\bibfield  {journal} {\bibinfo
  {journal} {Machine Learning with Applications}\ }\textbf {\bibinfo {volume}
  {6}},\ \bibinfo {pages} {100134} (\bibinfo {year} {2021})}\BibitemShut
  {NoStop}%
\bibitem [{\citenamefont {Minaee}\ \emph {et~al.}(2021)\citenamefont {Minaee},
  \citenamefont {Boykov}, \citenamefont {Porikli}, \citenamefont {Plaza},
  \citenamefont {Kehtarnavaz},\ and\ \citenamefont
  {Terzopoulos}}]{minaee_image_2021}%
  \BibitemOpen
  \bibfield  {author} {\bibinfo {author} {\bibfnamefont {S.}~\bibnamefont
  {Minaee}}, \bibinfo {author} {\bibfnamefont {Y.~Y.}\ \bibnamefont {Boykov}},
  \bibinfo {author} {\bibfnamefont {F.}~\bibnamefont {Porikli}}, \bibinfo
  {author} {\bibfnamefont {A.~J.}\ \bibnamefont {Plaza}}, \bibinfo {author}
  {\bibfnamefont {N.}~\bibnamefont {Kehtarnavaz}},\ and\ \bibinfo {author}
  {\bibfnamefont {D.}~\bibnamefont {Terzopoulos}},\ }\bibfield  {title}
  {\bibinfo {title} {Image {{Segmentation Using Deep Learning}}: {{A
  Survey}}},\ }\href {https://doi.org/10.1109/TPAMI.2021.3059968} {\bibfield
  {journal} {\bibinfo  {journal} {IEEE Transactions on Pattern Analysis and
  Machine Intelligence}\ ,\ \bibinfo {pages} {1}} (\bibinfo {year}
  {2021})}\BibitemShut {NoStop}%
\bibitem [{Note1()}]{Note1}%
  \BibitemOpen
  \bibinfo {note} {Https://mlcommons.org/en/}\BibitemShut {NoStop}%
\bibitem [{\citenamefont {Chollet}\ \emph {et~al.}(2015)\citenamefont {Chollet}
  \emph {et~al.}}]{chollet2015keras}%
  \BibitemOpen
  \bibfield  {author} {\bibinfo {author} {\bibfnamefont {F.}~\bibnamefont
  {Chollet}} \emph {et~al.},\ }\href@noop {} {\bibinfo {title} {Keras}}
  (\bibinfo {year} {2015})\BibitemShut {NoStop}%
\bibitem [{\citenamefont {Midtvedt}\ \emph {et~al.}(2021)\citenamefont
  {Midtvedt}, \citenamefont {Helgadottir}, \citenamefont {Argun}, \citenamefont
  {Pineda}, \citenamefont {Midtvedt},\ and\ \citenamefont
  {Volpe}}]{midtvedt_quantitative_2021}%
  \BibitemOpen
  \bibfield  {author} {\bibinfo {author} {\bibfnamefont {B.}~\bibnamefont
  {Midtvedt}}, \bibinfo {author} {\bibfnamefont {S.}~\bibnamefont
  {Helgadottir}}, \bibinfo {author} {\bibfnamefont {A.}~\bibnamefont {Argun}},
  \bibinfo {author} {\bibfnamefont {J.}~\bibnamefont {Pineda}}, \bibinfo
  {author} {\bibfnamefont {D.}~\bibnamefont {Midtvedt}},\ and\ \bibinfo
  {author} {\bibfnamefont {G.}~\bibnamefont {Volpe}},\ }\bibfield  {title}
  {\bibinfo {title} {Quantitative digital microscopy with deep learning},\
  }\href {https://doi.org/10.1063/5.0034891} {\bibfield  {journal} {\bibinfo
  {journal} {Applied Physics Reviews}\ }\textbf {\bibinfo {volume} {8}},\
  \bibinfo {pages} {011310} (\bibinfo {year} {2021})}\BibitemShut {NoStop}%
\bibitem [{\citenamefont {Ershov}\ \emph {et~al.}(2022)\citenamefont {Ershov},
  \citenamefont {Phan}, \citenamefont {Pylv{\"a}n{\"a}inen}, \citenamefont
  {Rigaud}, \citenamefont {Le~Blanc}, \citenamefont {{Charles-Orszag}},
  \citenamefont {Conway}, \citenamefont {Laine}, \citenamefont {Roy},
  \citenamefont {Bonazzi}, \citenamefont {Dum{\'e}nil}, \citenamefont
  {Jacquemet},\ and\ \citenamefont {Tinevez}}]{ershov_trackmate_2022}%
  \BibitemOpen
  \bibfield  {author} {\bibinfo {author} {\bibfnamefont {D.}~\bibnamefont
  {Ershov}}, \bibinfo {author} {\bibfnamefont {M.-S.}\ \bibnamefont {Phan}},
  \bibinfo {author} {\bibfnamefont {J.~W.}\ \bibnamefont
  {Pylv{\"a}n{\"a}inen}}, \bibinfo {author} {\bibfnamefont {S.~U.}\
  \bibnamefont {Rigaud}}, \bibinfo {author} {\bibfnamefont {L.}~\bibnamefont
  {Le~Blanc}}, \bibinfo {author} {\bibfnamefont {A.}~\bibnamefont
  {{Charles-Orszag}}}, \bibinfo {author} {\bibfnamefont {J.~R.~W.}\
  \bibnamefont {Conway}}, \bibinfo {author} {\bibfnamefont {R.~F.}\
  \bibnamefont {Laine}}, \bibinfo {author} {\bibfnamefont {N.~H.}\ \bibnamefont
  {Roy}}, \bibinfo {author} {\bibfnamefont {D.}~\bibnamefont {Bonazzi}},
  \bibinfo {author} {\bibfnamefont {G.}~\bibnamefont {Dum{\'e}nil}}, \bibinfo
  {author} {\bibfnamefont {G.}~\bibnamefont {Jacquemet}},\ and\ \bibinfo
  {author} {\bibfnamefont {J.-Y.}\ \bibnamefont {Tinevez}},\ }\bibfield
  {title} {\bibinfo {title} {{{TrackMate}} 7: Integrating state-of-the-art
  segmentation algorithms into tracking pipelines},\ }\href
  {https://doi.org/10.1038/s41592-022-01507-1} {\bibfield  {journal} {\bibinfo
  {journal} {Nature Methods}\ }\textbf {\bibinfo {volume} {19}},\ \bibinfo
  {pages} {829} (\bibinfo {year} {2022})}\BibitemShut {NoStop}%
\bibitem [{\citenamefont {Ulman}\ \emph {et~al.}(2017)\citenamefont {Ulman},
  \citenamefont {Ma{\v s}ka}, \citenamefont {Magnusson}, \citenamefont
  {Ronneberger}, \citenamefont {Haubold}, \citenamefont {Harder}, \citenamefont
  {Matula}, \citenamefont {Matula}, \citenamefont {Svoboda}, \citenamefont
  {Radojevic}, \citenamefont {Smal}, \citenamefont {Rohr}, \citenamefont
  {Jald{\'e}n}, \citenamefont {Blau}, \citenamefont {Dzyubachyk}, \citenamefont
  {Lelieveldt}, \citenamefont {Xiao}, \citenamefont {Li}, \citenamefont {Cho},
  \citenamefont {Dufour}, \citenamefont {{Olivo-Marin}}, \citenamefont
  {{Reyes-Aldasoro}}, \citenamefont {{Solis-Lemus}}, \citenamefont {Bensch},
  \citenamefont {Brox}, \citenamefont {Stegmaier}, \citenamefont {Mikut},
  \citenamefont {Wolf}, \citenamefont {Hamprecht}, \citenamefont {Esteves},
  \citenamefont {Quelhas}, \citenamefont {Demirel}, \citenamefont
  {Malmstr{\"o}m}, \citenamefont {Jug}, \citenamefont {Tomancak}, \citenamefont
  {Meijering}, \citenamefont {{Mu{\~n}oz-Barrutia}}, \citenamefont {Kozubek},\
  and\ \citenamefont {{Ortiz-de-Solorzano}}}]{ulman_objective_2017}%
  \BibitemOpen
  \bibfield  {author} {\bibinfo {author} {\bibfnamefont {V.}~\bibnamefont
  {Ulman}}, \bibinfo {author} {\bibfnamefont {M.}~\bibnamefont {Ma{\v s}ka}},
  \bibinfo {author} {\bibfnamefont {K.~E.~G.}\ \bibnamefont {Magnusson}},
  \bibinfo {author} {\bibfnamefont {O.}~\bibnamefont {Ronneberger}}, \bibinfo
  {author} {\bibfnamefont {C.}~\bibnamefont {Haubold}}, \bibinfo {author}
  {\bibfnamefont {N.}~\bibnamefont {Harder}}, \bibinfo {author} {\bibfnamefont
  {P.}~\bibnamefont {Matula}}, \bibinfo {author} {\bibfnamefont
  {P.}~\bibnamefont {Matula}}, \bibinfo {author} {\bibfnamefont
  {D.}~\bibnamefont {Svoboda}}, \bibinfo {author} {\bibfnamefont
  {M.}~\bibnamefont {Radojevic}}, \bibinfo {author} {\bibfnamefont
  {I.}~\bibnamefont {Smal}}, \bibinfo {author} {\bibfnamefont {K.}~\bibnamefont
  {Rohr}}, \bibinfo {author} {\bibfnamefont {J.}~\bibnamefont {Jald{\'e}n}},
  \bibinfo {author} {\bibfnamefont {H.~M.}\ \bibnamefont {Blau}}, \bibinfo
  {author} {\bibfnamefont {O.}~\bibnamefont {Dzyubachyk}}, \bibinfo {author}
  {\bibfnamefont {B.}~\bibnamefont {Lelieveldt}}, \bibinfo {author}
  {\bibfnamefont {P.}~\bibnamefont {Xiao}}, \bibinfo {author} {\bibfnamefont
  {Y.}~\bibnamefont {Li}}, \bibinfo {author} {\bibfnamefont {S.-Y.}\
  \bibnamefont {Cho}}, \bibinfo {author} {\bibfnamefont {A.~C.}\ \bibnamefont
  {Dufour}}, \bibinfo {author} {\bibfnamefont {J.-C.}\ \bibnamefont
  {{Olivo-Marin}}}, \bibinfo {author} {\bibfnamefont {C.~C.}\ \bibnamefont
  {{Reyes-Aldasoro}}}, \bibinfo {author} {\bibfnamefont {J.~A.}\ \bibnamefont
  {{Solis-Lemus}}}, \bibinfo {author} {\bibfnamefont {R.}~\bibnamefont
  {Bensch}}, \bibinfo {author} {\bibfnamefont {T.}~\bibnamefont {Brox}},
  \bibinfo {author} {\bibfnamefont {J.}~\bibnamefont {Stegmaier}}, \bibinfo
  {author} {\bibfnamefont {R.}~\bibnamefont {Mikut}}, \bibinfo {author}
  {\bibfnamefont {S.}~\bibnamefont {Wolf}}, \bibinfo {author} {\bibfnamefont
  {F.~A.}\ \bibnamefont {Hamprecht}}, \bibinfo {author} {\bibfnamefont
  {T.}~\bibnamefont {Esteves}}, \bibinfo {author} {\bibfnamefont
  {P.}~\bibnamefont {Quelhas}}, \bibinfo {author} {\bibfnamefont
  {{\"O}.}~\bibnamefont {Demirel}}, \bibinfo {author} {\bibfnamefont
  {L.}~\bibnamefont {Malmstr{\"o}m}}, \bibinfo {author} {\bibfnamefont
  {F.}~\bibnamefont {Jug}}, \bibinfo {author} {\bibfnamefont {P.}~\bibnamefont
  {Tomancak}}, \bibinfo {author} {\bibfnamefont {E.}~\bibnamefont {Meijering}},
  \bibinfo {author} {\bibfnamefont {A.}~\bibnamefont {{Mu{\~n}oz-Barrutia}}},
  \bibinfo {author} {\bibfnamefont {M.}~\bibnamefont {Kozubek}},\ and\ \bibinfo
  {author} {\bibfnamefont {C.}~\bibnamefont {{Ortiz-de-Solorzano}}},\
  }\bibfield  {title} {\bibinfo {title} {An objective comparison of
  cell-tracking algorithms},\ }\href {https://doi.org/10.1038/nmeth.4473}
  {\bibfield  {journal} {\bibinfo  {journal} {Nature Methods}\ }\textbf
  {\bibinfo {volume} {14}},\ \bibinfo {pages} {1141} (\bibinfo {year}
  {2017})}\BibitemShut {NoStop}%
\bibitem [{\citenamefont {Stringer}\ \emph {et~al.}(2021)\citenamefont
  {Stringer}, \citenamefont {Wang}, \citenamefont {Michaelos},\ and\
  \citenamefont {Pachitariu}}]{stringer_cellpose_2021}%
  \BibitemOpen
  \bibfield  {author} {\bibinfo {author} {\bibfnamefont {C.}~\bibnamefont
  {Stringer}}, \bibinfo {author} {\bibfnamefont {T.}~\bibnamefont {Wang}},
  \bibinfo {author} {\bibfnamefont {M.}~\bibnamefont {Michaelos}},\ and\
  \bibinfo {author} {\bibfnamefont {M.}~\bibnamefont {Pachitariu}},\ }\bibfield
   {title} {\bibinfo {title} {Cellpose: A generalist algorithm for cellular
  segmentation},\ }\href {https://doi.org/10.1038/s41592-020-01018-x}
  {\bibfield  {journal} {\bibinfo  {journal} {Nature Methods}\ }\textbf
  {\bibinfo {volume} {18}},\ \bibinfo {pages} {100} (\bibinfo {year}
  {2021})}\BibitemShut {NoStop}%
\bibitem [{\citenamefont {Zheng}\ \emph {et~al.}(2022)\citenamefont {Zheng},
  \citenamefont {Jackson}, \citenamefont {Fortier}, \citenamefont {Bonassar},
  \citenamefont {Delco},\ and\ \citenamefont {Cohen}}]{zheng_strains_2022}%
  \BibitemOpen
  \bibfield  {author} {\bibinfo {author} {\bibfnamefont {J.}~\bibnamefont
  {Zheng}}, \bibinfo {author} {\bibfnamefont {T.~W.}\ \bibnamefont {Jackson}},
  \bibinfo {author} {\bibfnamefont {L.~A.}\ \bibnamefont {Fortier}}, \bibinfo
  {author} {\bibfnamefont {L.~J.}\ \bibnamefont {Bonassar}}, \bibinfo {author}
  {\bibfnamefont {M.~L.}\ \bibnamefont {Delco}},\ and\ \bibinfo {author}
  {\bibfnamefont {I.}~\bibnamefont {Cohen}},\ }\bibfield  {title} {\bibinfo
  {title} {{{STRAINS}}: {{A}} big data method for classifying cellular response
  to stimuli at the tissue scale},\ }\href
  {https://doi.org/10.1371/journal.pone.0278626} {\bibfield  {journal}
  {\bibinfo  {journal} {PLOS ONE}\ }\textbf {\bibinfo {volume} {17}},\ \bibinfo
  {pages} {e0278626} (\bibinfo {year} {2022})}\BibitemShut {NoStop}%
\bibitem [{\citenamefont {Newby}\ \emph {et~al.}(2018)\citenamefont {Newby},
  \citenamefont {Schaefer}, \citenamefont {Lee}, \citenamefont {Forest},\ and\
  \citenamefont {Lai}}]{newby_convolutional_2018}%
  \BibitemOpen
  \bibfield  {author} {\bibinfo {author} {\bibfnamefont {J.~M.}\ \bibnamefont
  {Newby}}, \bibinfo {author} {\bibfnamefont {A.~M.}\ \bibnamefont {Schaefer}},
  \bibinfo {author} {\bibfnamefont {P.~T.}\ \bibnamefont {Lee}}, \bibinfo
  {author} {\bibfnamefont {M.~G.}\ \bibnamefont {Forest}},\ and\ \bibinfo
  {author} {\bibfnamefont {S.~K.}\ \bibnamefont {Lai}},\ }\bibfield  {title}
  {\bibinfo {title} {Convolutional neural networks automate detection for
  tracking of submicron-scale particles in {{2D}} and {{3D}}},\ }\href
  {https://doi.org/10.1073/pnas.1804420115} {\bibfield  {journal} {\bibinfo
  {journal} {Proceedings of the National Academy of Sciences}\ }\textbf
  {\bibinfo {volume} {115}},\ \bibinfo {pages} {9026} (\bibinfo {year}
  {2018})}\BibitemShut {NoStop}%
\bibitem [{\citenamefont {Berg}\ \emph {et~al.}(2019)\citenamefont {Berg},
  \citenamefont {Kutra}, \citenamefont {Kroeger}, \citenamefont {Straehle},
  \citenamefont {Kausler}, \citenamefont {Haubold}, \citenamefont {Schiegg},
  \citenamefont {Ales}, \citenamefont {Beier}, \citenamefont {Rudy},
  \citenamefont {Eren}, \citenamefont {Cervantes}, \citenamefont {Xu},
  \citenamefont {Beuttenmueller}, \citenamefont {Wolny}, \citenamefont {Zhang},
  \citenamefont {Koethe}, \citenamefont {Hamprecht},\ and\ \citenamefont
  {Kreshuk}}]{berg_ilastik_2019}%
  \BibitemOpen
  \bibfield  {author} {\bibinfo {author} {\bibfnamefont {S.}~\bibnamefont
  {Berg}}, \bibinfo {author} {\bibfnamefont {D.}~\bibnamefont {Kutra}},
  \bibinfo {author} {\bibfnamefont {T.}~\bibnamefont {Kroeger}}, \bibinfo
  {author} {\bibfnamefont {C.~N.}\ \bibnamefont {Straehle}}, \bibinfo {author}
  {\bibfnamefont {B.~X.}\ \bibnamefont {Kausler}}, \bibinfo {author}
  {\bibfnamefont {C.}~\bibnamefont {Haubold}}, \bibinfo {author} {\bibfnamefont
  {M.}~\bibnamefont {Schiegg}}, \bibinfo {author} {\bibfnamefont
  {J.}~\bibnamefont {Ales}}, \bibinfo {author} {\bibfnamefont {T.}~\bibnamefont
  {Beier}}, \bibinfo {author} {\bibfnamefont {M.}~\bibnamefont {Rudy}},
  \bibinfo {author} {\bibfnamefont {K.}~\bibnamefont {Eren}}, \bibinfo {author}
  {\bibfnamefont {J.~I.}\ \bibnamefont {Cervantes}}, \bibinfo {author}
  {\bibfnamefont {B.}~\bibnamefont {Xu}}, \bibinfo {author} {\bibfnamefont
  {F.}~\bibnamefont {Beuttenmueller}}, \bibinfo {author} {\bibfnamefont
  {A.}~\bibnamefont {Wolny}}, \bibinfo {author} {\bibfnamefont
  {C.}~\bibnamefont {Zhang}}, \bibinfo {author} {\bibfnamefont
  {U.}~\bibnamefont {Koethe}}, \bibinfo {author} {\bibfnamefont {F.~A.}\
  \bibnamefont {Hamprecht}},\ and\ \bibinfo {author} {\bibfnamefont
  {A.}~\bibnamefont {Kreshuk}},\ }\bibfield  {title} {\bibinfo {title}
  {Ilastik: Interactive machine learning for (bio)image analysis},\ }\href
  {https://doi.org/10.1038/s41592-019-0582-9} {\bibfield  {journal} {\bibinfo
  {journal} {Nature Methods}\ }\textbf {\bibinfo {volume} {16}},\ \bibinfo
  {pages} {1226} (\bibinfo {year} {2019})}\BibitemShut {NoStop}%
\bibitem [{\citenamefont {Nguyen}\ \emph {et~al.}(2017)\citenamefont {Nguyen},
  \citenamefont {Linder}, \citenamefont {Plummer}, \citenamefont {Shaevitz},\
  and\ \citenamefont {Leifer}}]{nguyen_automatically_2017}%
  \BibitemOpen
  \bibfield  {author} {\bibinfo {author} {\bibfnamefont {J.~P.}\ \bibnamefont
  {Nguyen}}, \bibinfo {author} {\bibfnamefont {A.~N.}\ \bibnamefont {Linder}},
  \bibinfo {author} {\bibfnamefont {G.~S.}\ \bibnamefont {Plummer}}, \bibinfo
  {author} {\bibfnamefont {J.~W.}\ \bibnamefont {Shaevitz}},\ and\ \bibinfo
  {author} {\bibfnamefont {A.~M.}\ \bibnamefont {Leifer}},\ }\bibfield  {title}
  {\bibinfo {title} {Automatically tracking neurons in a moving and deforming
  brain},\ }\href {https://doi.org/10.1371/journal.pcbi.1005517} {\bibfield
  {journal} {\bibinfo  {journal} {PLOS Computational Biology}\ }\textbf
  {\bibinfo {volume} {13}},\ \bibinfo {pages} {e1005517} (\bibinfo {year}
  {2017})}\BibitemShut {NoStop}%
\bibitem [{\citenamefont {Masubuchi}\ \emph {et~al.}(2020)\citenamefont
  {Masubuchi}, \citenamefont {Watanabe}, \citenamefont {Seo}, \citenamefont
  {Okazaki}, \citenamefont {Sasagawa}, \citenamefont {Watanabe}, \citenamefont
  {Taniguchi},\ and\ \citenamefont
  {Machida}}]{masubuchi_deeplearningbased_2020}%
  \BibitemOpen
  \bibfield  {author} {\bibinfo {author} {\bibfnamefont {S.}~\bibnamefont
  {Masubuchi}}, \bibinfo {author} {\bibfnamefont {E.}~\bibnamefont {Watanabe}},
  \bibinfo {author} {\bibfnamefont {Y.}~\bibnamefont {Seo}}, \bibinfo {author}
  {\bibfnamefont {S.}~\bibnamefont {Okazaki}}, \bibinfo {author} {\bibfnamefont
  {T.}~\bibnamefont {Sasagawa}}, \bibinfo {author} {\bibfnamefont
  {K.}~\bibnamefont {Watanabe}}, \bibinfo {author} {\bibfnamefont
  {T.}~\bibnamefont {Taniguchi}},\ and\ \bibinfo {author} {\bibfnamefont
  {T.}~\bibnamefont {Machida}},\ }\bibfield  {title} {\bibinfo {title}
  {Deep-learning-based image segmentation integrated with optical microscopy
  for automatically searching for two-dimensional materials},\ }\href
  {https://doi.org/10.1038/s41699-020-0137-z} {\bibfield  {journal} {\bibinfo
  {journal} {npj 2D Materials and Applications}\ }\textbf {\bibinfo {volume}
  {4}},\ \bibinfo {pages} {1} (\bibinfo {year} {2020})}\BibitemShut {NoStop}%
\bibitem [{\citenamefont {Malhotra}\ \emph {et~al.}(2022)\citenamefont
  {Malhotra}, \citenamefont {Gupta}, \citenamefont {Koundal}, \citenamefont
  {Zaguia},\ and\ \citenamefont {Enbeyle}}]{malhotra_deep_2022}%
  \BibitemOpen
  \bibfield  {author} {\bibinfo {author} {\bibfnamefont {P.}~\bibnamefont
  {Malhotra}}, \bibinfo {author} {\bibfnamefont {S.}~\bibnamefont {Gupta}},
  \bibinfo {author} {\bibfnamefont {D.}~\bibnamefont {Koundal}}, \bibinfo
  {author} {\bibfnamefont {A.}~\bibnamefont {Zaguia}},\ and\ \bibinfo {author}
  {\bibfnamefont {W.}~\bibnamefont {Enbeyle}},\ }\bibfield  {title} {\bibinfo
  {title} {Deep {{Neural Networks}} for {{Medical Image Segmentation}}},\
  }\href {https://doi.org/10.1155/2022/9580991} {\bibfield  {journal} {\bibinfo
   {journal} {Journal of Healthcare Engineering}\ }\textbf {\bibinfo {volume}
  {2022}},\ \bibinfo {pages} {e9580991} (\bibinfo {year} {2022})}\BibitemShut
  {NoStop}%
\bibitem [{\citenamefont {Ronneberger}\ \emph {et~al.}(2015)\citenamefont
  {Ronneberger}, \citenamefont {Fischer},\ and\ \citenamefont
  {Brox}}]{ronneberger_unet_2015}%
  \BibitemOpen
  \bibfield  {author} {\bibinfo {author} {\bibfnamefont {O.}~\bibnamefont
  {Ronneberger}}, \bibinfo {author} {\bibfnamefont {P.}~\bibnamefont
  {Fischer}},\ and\ \bibinfo {author} {\bibfnamefont {T.}~\bibnamefont
  {Brox}},\ }\bibfield  {title} {\bibinfo {title} {U-{{Net}}: {{Convolutional
  Networks}} for {{Biomedical Image Segmentation}}},\ }in\ \href
  {https://doi.org/10.1007/978-3-319-24574-4_28} {\emph {\bibinfo {booktitle}
  {Medical {{Image Computing}} and {{Computer-Assisted Intervention}}
  \textendash{} {{MICCAI}} 2015}}},\ Vol.\ \bibinfo {volume} {9351},\ \bibinfo
  {editor} {edited by\ \bibinfo {editor} {\bibfnamefont {N.}~\bibnamefont
  {Navab}}, \bibinfo {editor} {\bibfnamefont {J.}~\bibnamefont {Hornegger}},
  \bibinfo {editor} {\bibfnamefont {W.~M.}\ \bibnamefont {Wells}},\ and\
  \bibinfo {editor} {\bibfnamefont {A.~F.}\ \bibnamefont {Frangi}}}\ (\bibinfo
  {publisher} {{Springer International Publishing}},\ \bibinfo {address}
  {{Cham}},\ \bibinfo {year} {2015})\ pp.\ \bibinfo {pages}
  {234--241}\BibitemShut {NoStop}%
\bibitem [{\citenamefont {Ciresan}\ \emph {et~al.}(2012)\citenamefont
  {Ciresan}, \citenamefont {Giusti}, \citenamefont {Gambardella},\ and\
  \citenamefont {Schmidhuber}}]{ciresan_deep_2012}%
  \BibitemOpen
  \bibfield  {author} {\bibinfo {author} {\bibfnamefont {D.}~\bibnamefont
  {Ciresan}}, \bibinfo {author} {\bibfnamefont {A.}~\bibnamefont {Giusti}},
  \bibinfo {author} {\bibfnamefont {L.}~\bibnamefont {Gambardella}},\ and\
  \bibinfo {author} {\bibfnamefont {J.}~\bibnamefont {Schmidhuber}},\
  }\bibfield  {title} {\bibinfo {title} {Deep {{Neural Networks Segment
  Neuronal Membranes}} in {{Electron Microscopy Images}}},\ }in\ \href@noop {}
  {\emph {\bibinfo {booktitle} {Advances in {{Neural Information Processing
  Systems}}}}},\ Vol.~\bibinfo {volume} {25}\ (\bibinfo  {publisher} {{Curran
  Associates, Inc.}},\ \bibinfo {year} {2012})\BibitemShut {NoStop}%
\bibitem [{\citenamefont {Haertter}\ \emph {et~al.}(2022)\citenamefont
  {Haertter}, \citenamefont {Wang}, \citenamefont {Fogerson}, \citenamefont
  {Ramkumar}, \citenamefont {Crawford}, \citenamefont {Poss}, \citenamefont
  {Di~Talia}, \citenamefont {Kiehart},\ and\ \citenamefont
  {Schmidt}}]{haertter_deepprojection_2022}%
  \BibitemOpen
  \bibfield  {author} {\bibinfo {author} {\bibfnamefont {D.}~\bibnamefont
  {Haertter}}, \bibinfo {author} {\bibfnamefont {X.}~\bibnamefont {Wang}},
  \bibinfo {author} {\bibfnamefont {S.~M.}\ \bibnamefont {Fogerson}}, \bibinfo
  {author} {\bibfnamefont {N.}~\bibnamefont {Ramkumar}}, \bibinfo {author}
  {\bibfnamefont {J.~M.}\ \bibnamefont {Crawford}}, \bibinfo {author}
  {\bibfnamefont {K.~D.}\ \bibnamefont {Poss}}, \bibinfo {author}
  {\bibfnamefont {S.}~\bibnamefont {Di~Talia}}, \bibinfo {author}
  {\bibfnamefont {D.~P.}\ \bibnamefont {Kiehart}},\ and\ \bibinfo {author}
  {\bibfnamefont {C.~F.}\ \bibnamefont {Schmidt}},\ }\bibfield  {title}
  {\bibinfo {title} {{{DeepProjection}}: Specific and robust projection of
  curved {{2D}} tissue sheets from {{3D}} microscopy using deep learning},\
  }\href {https://doi.org/10.1242/dev.200621} {\bibfield  {journal} {\bibinfo
  {journal} {Development}\ }\textbf {\bibinfo {volume} {149}},\ \bibinfo
  {pages} {dev200621} (\bibinfo {year} {2022})}\BibitemShut {NoStop}%
\bibitem [{\citenamefont {Niedballa}\ \emph {et~al.}(2022)\citenamefont
  {Niedballa}, \citenamefont {Axtner}, \citenamefont {D{\"o}bert},
  \citenamefont {Tilker}, \citenamefont {Nguyen}, \citenamefont {Wong},
  \citenamefont {Fiderer}, \citenamefont {Heurich},\ and\ \citenamefont
  {Wilting}}]{niedballa_imageseg_2022}%
  \BibitemOpen
  \bibfield  {author} {\bibinfo {author} {\bibfnamefont {J.}~\bibnamefont
  {Niedballa}}, \bibinfo {author} {\bibfnamefont {J.}~\bibnamefont {Axtner}},
  \bibinfo {author} {\bibfnamefont {T.~F.}\ \bibnamefont {D{\"o}bert}},
  \bibinfo {author} {\bibfnamefont {A.}~\bibnamefont {Tilker}}, \bibinfo
  {author} {\bibfnamefont {A.}~\bibnamefont {Nguyen}}, \bibinfo {author}
  {\bibfnamefont {S.~T.}\ \bibnamefont {Wong}}, \bibinfo {author}
  {\bibfnamefont {C.}~\bibnamefont {Fiderer}}, \bibinfo {author} {\bibfnamefont
  {M.}~\bibnamefont {Heurich}},\ and\ \bibinfo {author} {\bibfnamefont
  {A.}~\bibnamefont {Wilting}},\ }\bibfield  {title} {\bibinfo {title}
  {Imageseg: {{An R}} package for deep learning-based image segmentation},\
  }\href {https://doi.org/10.1111/2041-210X.13984} {\bibfield  {journal}
  {\bibinfo  {journal} {Methods in Ecology and Evolution}\ }\textbf {\bibinfo
  {volume} {13}},\ \bibinfo {pages} {2363} (\bibinfo {year}
  {2022})}\BibitemShut {NoStop}%
\bibitem [{\citenamefont {R{\"u}hle}\ \emph {et~al.}(2021)\citenamefont
  {R{\"u}hle}, \citenamefont {Krumrey},\ and\ \citenamefont
  {Hodoroaba}}]{ruhle_workflow_2021}%
  \BibitemOpen
  \bibfield  {author} {\bibinfo {author} {\bibfnamefont {B.}~\bibnamefont
  {R{\"u}hle}}, \bibinfo {author} {\bibfnamefont {J.~F.}\ \bibnamefont
  {Krumrey}},\ and\ \bibinfo {author} {\bibfnamefont {V.-D.}\ \bibnamefont
  {Hodoroaba}},\ }\bibfield  {title} {\bibinfo {title} {Workflow towards
  automated segmentation of agglomerated, non-spherical particles from electron
  microscopy images using artificial neural networks},\ }\href
  {https://doi.org/10.1038/s41598-021-84287-6} {\bibfield  {journal} {\bibinfo
  {journal} {Scientific Reports}\ }\textbf {\bibinfo {volume} {11}},\ \bibinfo
  {pages} {4942} (\bibinfo {year} {2021})}\BibitemShut {NoStop}%
\bibitem [{\citenamefont {Azimi}\ \emph {et~al.}(2018)\citenamefont {Azimi},
  \citenamefont {Britz}, \citenamefont {Engstler}, \citenamefont {Fritz},\ and\
  \citenamefont {M{\"u}cklich}}]{azimi_advanced_2018}%
  \BibitemOpen
  \bibfield  {author} {\bibinfo {author} {\bibfnamefont {S.~M.}\ \bibnamefont
  {Azimi}}, \bibinfo {author} {\bibfnamefont {D.}~\bibnamefont {Britz}},
  \bibinfo {author} {\bibfnamefont {M.}~\bibnamefont {Engstler}}, \bibinfo
  {author} {\bibfnamefont {M.}~\bibnamefont {Fritz}},\ and\ \bibinfo {author}
  {\bibfnamefont {F.}~\bibnamefont {M{\"u}cklich}},\ }\bibfield  {title}
  {\bibinfo {title} {Advanced {{Steel Microstructural Classification}} by
  {{Deep Learning Methods}}},\ }\href
  {https://doi.org/10.1038/s41598-018-20037-5} {\bibfield  {journal} {\bibinfo
  {journal} {Scientific Reports}\ }\textbf {\bibinfo {volume} {8}},\ \bibinfo
  {pages} {2128} (\bibinfo {year} {2018})}\BibitemShut {NoStop}%
\bibitem [{\citenamefont {Ostdiek}\ \emph {et~al.}(2022)\citenamefont
  {Ostdiek}, \citenamefont {Diaz~Rivero},\ and\ \citenamefont
  {Dvorkin}}]{ostdiek_image_2022}%
  \BibitemOpen
  \bibfield  {author} {\bibinfo {author} {\bibfnamefont {B.}~\bibnamefont
  {Ostdiek}}, \bibinfo {author} {\bibfnamefont {A.}~\bibnamefont
  {Diaz~Rivero}},\ and\ \bibinfo {author} {\bibfnamefont {C.}~\bibnamefont
  {Dvorkin}},\ }\bibfield  {title} {\bibinfo {title} {Image segmentation for
  analyzing galaxy-galaxy strong lensing systems},\ }\href
  {https://doi.org/10.1051/0004-6361/202142030} {\bibfield  {journal} {\bibinfo
   {journal} {Astronomy \& Astrophysics}\ }\textbf {\bibinfo {volume} {657}},\
  \bibinfo {pages} {L14} (\bibinfo {year} {2022})}\BibitemShut {NoStop}%
\bibitem [{\citenamefont {Hausen}\ and\ \citenamefont
  {Robertson}(2020)}]{hausen_morpheus_2020}%
  \BibitemOpen
  \bibfield  {author} {\bibinfo {author} {\bibfnamefont {R.}~\bibnamefont
  {Hausen}}\ and\ \bibinfo {author} {\bibfnamefont {B.~E.}\ \bibnamefont
  {Robertson}},\ }\bibfield  {title} {\bibinfo {title} {Morpheus: {{A Deep
  Learning Framework}} for the {{Pixel-level Analysis}} of {{Astronomical Image
  Data}}},\ }\href {https://doi.org/10.3847/1538-4365/ab8868} {\bibfield
  {journal} {\bibinfo  {journal} {The Astrophysical Journal Supplement Series}\
  }\textbf {\bibinfo {volume} {248}},\ \bibinfo {pages} {20} (\bibinfo {year}
  {2020})}\BibitemShut {NoStop}%
\bibitem [{\citenamefont {Li}\ \emph {et~al.}(2021)\citenamefont {Li},
  \citenamefont {Li},\ and\ \citenamefont {Xu}}]{li_reconstructing_2021}%
  \BibitemOpen
  \bibfield  {author} {\bibinfo {author} {\bibfnamefont {J.}~\bibnamefont
  {Li}}, \bibinfo {author} {\bibfnamefont {T.}~\bibnamefont {Li}},\ and\
  \bibinfo {author} {\bibfnamefont {F.-Z.}\ \bibnamefont {Xu}},\ }\bibfield
  {title} {\bibinfo {title} {Reconstructing boosted {{Higgs}} jets from event
  image segmentation},\ }\href {https://doi.org/10.1007/JHEP04(2021)156}
  {\bibfield  {journal} {\bibinfo  {journal} {Journal of High Energy Physics}\
  }\textbf {\bibinfo {volume} {2021}},\ \bibinfo {pages} {156} (\bibinfo {year}
  {2021})},\ \Eprint {https://arxiv.org/abs/2008.13529} {arxiv:2008.13529
  [hep-ex, physics:hep-ph]} \BibitemShut {NoStop}%
\bibitem [{\citenamefont {He}\ \emph {et~al.}(2015)\citenamefont {He},
  \citenamefont {Zhang}, \citenamefont {Ren},\ and\ \citenamefont
  {Sun}}]{he_deep_2015}%
  \BibitemOpen
  \bibfield  {author} {\bibinfo {author} {\bibfnamefont {K.}~\bibnamefont
  {He}}, \bibinfo {author} {\bibfnamefont {X.}~\bibnamefont {Zhang}}, \bibinfo
  {author} {\bibfnamefont {S.}~\bibnamefont {Ren}},\ and\ \bibinfo {author}
  {\bibfnamefont {J.}~\bibnamefont {Sun}},\ }\bibfield  {title} {\bibinfo
  {title} {Deep {{Residual Learning}} for {{Image Recognition}}},\ }\href@noop
  {} {\bibfield  {journal} {\bibinfo  {journal} {arXiv:1512.03385 [cs]}\ }
  (\bibinfo {year} {2015})},\ \Eprint {https://arxiv.org/abs/1512.03385}
  {arxiv:1512.03385 [cs]} \BibitemShut {NoStop}%
\bibitem [{\citenamefont {Pasquet}\ \emph {et~al.}(2023)\citenamefont
  {Pasquet}, \citenamefont {Galvani}, \citenamefont {Pitois}, \citenamefont
  {{Cohen-Addad}}, \citenamefont {H{\"o}hler}, \citenamefont {Chieco},
  \citenamefont {Dillavou}, \citenamefont {Hanlan}, \citenamefont {Durian},
  \citenamefont {Rio}, \citenamefont {Salonen},\ and\ \citenamefont
  {Langevin}}]{pasquet_aqueous_2023}%
  \BibitemOpen
  \bibfield  {author} {\bibinfo {author} {\bibfnamefont {M.}~\bibnamefont
  {Pasquet}}, \bibinfo {author} {\bibfnamefont {N.}~\bibnamefont {Galvani}},
  \bibinfo {author} {\bibfnamefont {O.}~\bibnamefont {Pitois}}, \bibinfo
  {author} {\bibfnamefont {S.}~\bibnamefont {{Cohen-Addad}}}, \bibinfo {author}
  {\bibfnamefont {R.}~\bibnamefont {H{\"o}hler}}, \bibinfo {author}
  {\bibfnamefont {A.~T.}\ \bibnamefont {Chieco}}, \bibinfo {author}
  {\bibfnamefont {S.}~\bibnamefont {Dillavou}}, \bibinfo {author}
  {\bibfnamefont {J.~M.}\ \bibnamefont {Hanlan}}, \bibinfo {author}
  {\bibfnamefont {D.~J.}\ \bibnamefont {Durian}}, \bibinfo {author}
  {\bibfnamefont {E.}~\bibnamefont {Rio}}, \bibinfo {author} {\bibfnamefont
  {A.}~\bibnamefont {Salonen}},\ and\ \bibinfo {author} {\bibfnamefont
  {D.}~\bibnamefont {Langevin}},\ }\bibfield  {title} {\bibinfo {title}
  {Aqueous foams in microgravity, measuring bubble sizes},\ }\href
  {https://doi.org/10.5802/crmeca.153} {\bibfield  {journal} {\bibinfo
  {journal} {Comptes Rendus. M\'ecanique}\ }\textbf {\bibinfo {volume} {351}},\
  \bibinfo {pages} {1} (\bibinfo {year} {2023})}\BibitemShut {NoStop}%
\bibitem [{\citenamefont {Daniels}\ \emph {et~al.}(2017)\citenamefont
  {Daniels}, \citenamefont {Kollmer},\ and\ \citenamefont
  {Puckett}}]{daniels_photoelastic_2017}%
  \BibitemOpen
  \bibfield  {author} {\bibinfo {author} {\bibfnamefont {K.~E.}\ \bibnamefont
  {Daniels}}, \bibinfo {author} {\bibfnamefont {J.~E.}\ \bibnamefont
  {Kollmer}},\ and\ \bibinfo {author} {\bibfnamefont {J.~G.}\ \bibnamefont
  {Puckett}},\ }\bibfield  {title} {\bibinfo {title} {Photoelastic force
  measurements in granular materials},\ }\href
  {https://doi.org/10.1063/1.4983049} {\bibfield  {journal} {\bibinfo
  {journal} {Review of Scientific Instruments}\ }\textbf {\bibinfo {volume}
  {88}},\ \bibinfo {pages} {051808} (\bibinfo {year} {2017})}\BibitemShut
  {NoStop}%
\bibitem [{Note2()}]{Note2}%
  \BibitemOpen
  \bibinfo {note}
  {Https://public.celltrackingchallenge.net/documents/SEG.pdf}\BibitemShut
  {NoStop}%
\bibitem [{Note3()}]{Note3}%
  \BibitemOpen
  \bibinfo {note} {This can also be modified easily via the parameters of the
  algorithm, to instead represent innies and outies in the ratio they are
  present in the images.}\BibitemShut {Stop}%
\end{thebibliography}%

\end{document}